\documentclass[a4paper,11pt]{article}
\usepackage{coling2018}
\usepackage{times}
\usepackage{url}
\usepackage{latexsym}
\usepackage{array}
\usepackage[]{algorithm}
\usepackage[noend]{algpseudocode}
\usepackage{ dsfont }

\makeatletter
\def\BState{\State\hskip-\ALG@thistlm}
\makeatother

\usepackage{microtype}

\usepackage{subfigure}
\usepackage{booktabs} 
\usepackage{times}
\usepackage{xcolor}
\usepackage{soul}
\usepackage[utf8]{inputenc}
\usepackage{epsfig}
\usepackage{graphicx}
\usepackage{amsmath}
\usepackage{amssymb}

\title{Learning Semantic Sentence Embeddings using Pair-wise Discriminator}

  \author{ $\textbf{Badri N. Patro}^*$  \quad $\textbf{Vinod K. Kurmi}^* $ \quad $\textbf{Sandeep Kumar}^*$ \quad  $\textbf{Vinay P. Namboodiri}$ \\
  Indian Institute of Technology, Kanpur \\
  {\tt \{badri,vinodkk,sandepkr,vinaypn\}@iitk.ac.in} \\
  \\}

\date{}

\begin{document}
\maketitle
\begin{abstract}
In this paper, we propose a method for obtaining sentence-level embeddings. While the problem of securing word-level embeddings is very well studied, we propose a novel method for obtaining sentence-level embeddings. This is obtained by a simple method in the context of solving the paraphrase generation task.  If we use a sequential encoder-decoder model for generating paraphrase, we would like the generated paraphrase to be semantically close to the original sentence. One way to ensure this is by adding constraints for true paraphrase embeddings to be close and unrelated paraphrase candidate sentence embeddings to be far. This is ensured by using a sequential pair-wise discriminator that shares weights with the encoder that is trained with a suitable loss function. Our loss function penalizes paraphrase sentence embedding distances from being too large. This loss is used in combination with a  sequential encoder-decoder network. We also validated our method by evaluating the obtained embeddings for a sentiment analysis task. The proposed method results in semantic embeddings and outperforms the state-of-the-art on the paraphrase generation and sentiment analysis task on standard datasets. These results are also shown to be statistically significant.
\end{abstract}

\section{Introduction}

\blfootnote{
    \hspace{-0.65cm}  
    * Equal contribution
}
 \blfootnote{
    \hspace{-0.65cm}  
    This work is licensed under a Creative Commons 
    Attribution 4.0 International License.
    License details:
    \url{http://creativecommons.org/licenses/by/4.0/}
}

The problem of obtaining a semantic embedding for a sentence that ensures that the related sentences are closer and unrelated sentences are farther lies at the core of understanding languages. This would be relevant for a wide variety of machine reading comprehension and related tasks such as sentiment analysis. Towards this problem, we propose a supervised method that uses a sequential encoder-decoder framework for paraphrase generation. The task of generating paraphrases is closely related to the task of obtaining semantic sentence embeddings. In our approach, we aim to ensure that the generated paraphrase embedding should be close to the true corresponding sentence and far from unrelated sentences. The embeddings so obtained help us to obtain state-of-the-art results for paraphrase generation task. 

\begin{figure*}[ht]
\centering
\includegraphics[width=0.95\columnwidth]{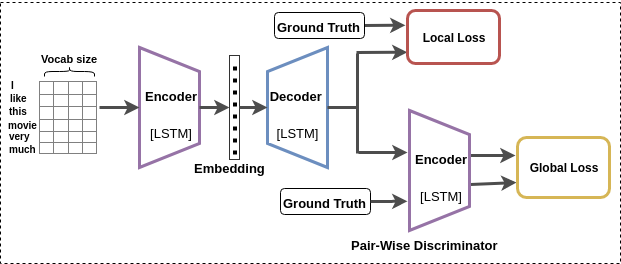} 
\caption{Pairwise Discriminator based Encoder-Decoder for Paraphrase Generation: This is the basic outline of our model which consists of an LSTM encoder, decoder and discriminator. Here the encoders share the weights. The discriminator generates discriminative embeddings for the Ground Truth-Generated paraphrase pair with the help of `global' loss. Our model is jointly trained with the help of a `local' and `global' loss which we describe in section~\ref{methods}.}
\label{fig:intro_fig}
\end{figure*}

Our model consists of a sequential encoder-decoder that is further trained using a pairwise discriminator. The encoder-decoder architecture has been widely used for machine translation and machine comprehension tasks. In general, the model ensures a `local' loss that is incurred for each recurrent unit cell. It only ensures that a particular word token is present at an appropriate place. This, however, does not imply that the whole sentence is correctly generated. To ensure that the whole sentence is correctly encoded, we make further use of a pair-wise discriminator that encodes the whole sentence and obtains an embedding for it. We further ensure that this is close to the desired ground-truth embeddings while being far from other (sentences in the corpus) embeddings. This model thus provides a `global' loss that ensures the sentence embedding as a whole is close to other semantically related sentence embeddings. This is illustrated in Figure~\ref{fig:intro_fig}. We further evaluate the validity of the sentence embeddings by using them for the task of sentiment analysis. We observe that the proposed sentence embeddings result in state-of-the-art performance for both these tasks.

Our contributions are: a) We propose a model for obtaining sentence embeddings for solving the paraphrase generation task using a pair-wise discriminator loss added to an encoder-decoder network. b) We show that these embeddings can also be used for the sentiment analysis task. c) We validate the model using standard datasets with a detailed comparison with state-of-the-art methods and also ensure that the results are statistically significant.

\section{Related Work}
Given the flexibility and diversity of natural language, it has always been a challenging task to represent text efficiently. There have been several hypotheses proposed for representing the same.~\cite{harris_TF1954,Firth_SLA1957,Sahlgren_IJDS2008} proposed a distribution hypothesis to represent words, i.e., words which occur in the same context have similar meanings. One popular hypothesis is the bag-of-words (BOW) or Vector Space Model~\cite{Salton_ACM1975}, in which a text (such as a sentence or a document) is represented as the bag (multiset) of its words.~\cite{lin_ACM2001dirt} proposed an extended distributional hypothesis and~\cite{Deerwester_JASIS1990,Turney_ACM2003} proposed a latent relation hypothesis, in which a pair of words that co-occur in similar patterns tend to have similar semantic relation.
Word2Vec\cite{Mikolov_ARXIV2013,mikolov_NIPS2013,Goldberg_ARXIV2014} is also a popular method for representing every unique word in the corpus in a vector space. Here, the embedding of every word is predicted based on its context (surrounding words). 
NLP researchers have also proposed phrase-level and sentence-level representations~\cite{mitchell_WOL2010,zanzotto_ACL2010,yessenalina_EMNLP2011,grefenstette_IWCS2013,mikolov_NIPS2013}.~\cite{socher_ICML2011,Kim_ARXIV2014,lin_EMNLP2015,Yin_ARXIV2015,Kalchbrenner_ARXIV2014} have analyzed several approaches to represent sentences and phrases by a weighted average of all the words in the sentence, combining the word vectors in an order given by a parse tree of a sentence and by using matrix-vector operations. The major issue with BOW models and weighted averaging of word vectors is the loss of semantic meaning of the words, the parse tree approaches can only work for sentences because of its dependence on sentence parsing mechanism.~\cite{Socher_EMNLP2013,Le_ICML2014} proposed a method to obtain a vector representation for paragraphs and use it to for some text-understanding problems like sentiment analysis and information retrieval. 

Many language models have been proposed for obtaining better text embeddings in Machine Translation~\cite{Sutskever_NIPS2014,cho_EMLNLP2014,vinyals_arXiv2015,wu_CoRR2016}, question generation~\cite{Du_ArXiv2017}, dialogue generation ~\cite{shang_ICNLP2015,li_ARXIV2016,li_arxiv2017adversarial}, document summarization~\cite{rush_EMNLP2015}, text  generation ~\cite{zhang_arxiv2017adversarial,Hu_ICML2017toward,Yu_AAAI2017seqgan,Guo_arxiv2017long,liang_CORR2017recurrent,Reed_CVPR2016} and question answering~\cite{yin_IJCAI2016,miao_ICML2016}. For paraphrase generation task,~\cite{Prakash_Arxiv2016neural} have generated paraphrases using stacked residual LSTM based network.~\cite{hasan_ClinicalNLP2016} proposed a encoder-decoder framework for this task.~\cite{gupta_AAAI2017} explored a VAE approach to  generate paraphrase sentences using recurrent neural networks.~\cite{li_arXiv2017Paraphrase} used reinforcement learning for paraphrase generation task. 


\label{sec:lit_surv}

\section{Method}\label{methods}
In this paper, we propose a text representation method for sentences based on an encoder-decoder framework using a pairwise discriminator for paraphrase generation and then fine tune these embeddings for sentiment analysis task. Our model is an extension of \textit{seq2seq}~\cite{Sutskever_NIPS2014} model for learning better text embeddings.
\subsection{Overview}
\begin{figure*}[ht]
\centering
\includegraphics[width=1.0\columnwidth]{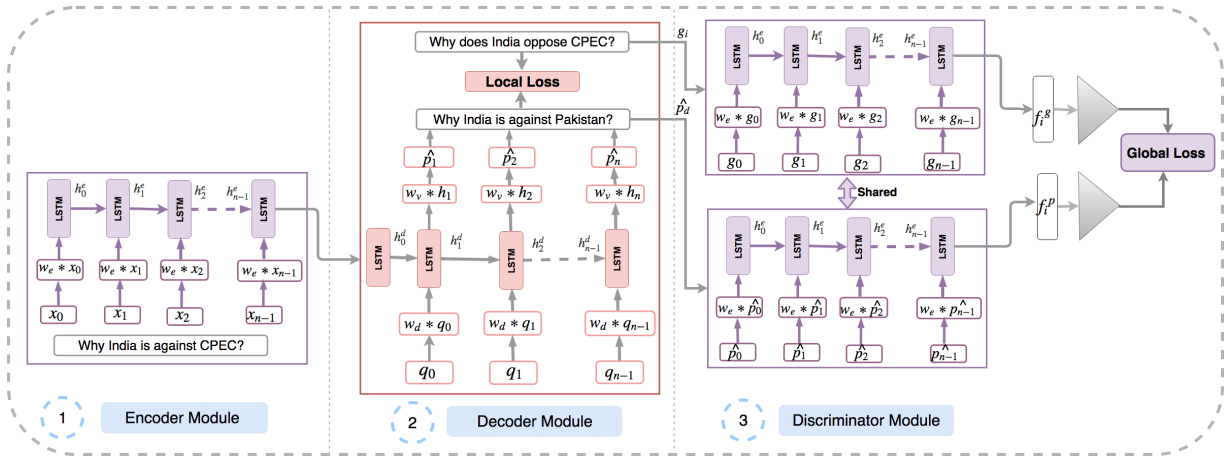} 
\caption{This is an overview of our model. It consists of 3 parts: 1) LSTM-based Encoder module which encodes a given sentence, 2) LSTM-based Decoder Module which generates natural language paraphrases from the encoded embeddings and 3) LSTM-based pairwise Discriminator module which shares its weights with the Encoder module and this whole network is trained with local and global loss.}
\label{fig:main_figure}
\end{figure*}

\noindent \textbf{Task:} In the paraphrase generation problem, given an input sequence of words  $X = [x_1,..., x_L]$, we need to generate another output sequence of words $Y = [q_1,..., q_T ]$ that has the same meaning as $X$. Here $L$ and $T$ are not fixed constants. Our training data consists of $M$ pairs of paraphrases  $ \{(X_i,Y_i)\}_{i=1}^M$ where $X_i$ and $Y_i$ are the paraphrase of each other.\\ \\
Our method consists of three modules as illustrated in Figure~\ref{fig:main_figure}: first is a Text Encoder which consists of LSTM layers, second is LSTM-based Text Decoder and last one is an LSTM-based Discriminator module. These are shown respectively in part 1, 2, 3 of Figure~\ref{fig:main_figure}. Our network with all three parts is trained end-to-end. The weight parameters of encoder and discriminator modules are shared. 
Instead of taking a separate discriminator, we shared it with the encoder so that it learns the embedding based on the `global' as well as `local' loss. After training, at test time we used encoder to generate feature maps and pass it to the decoder for generating paraphrases. These text embeddings can be further used for other NLP tasks such as sentiment analysis.

\subsection{Encoder-LSTM}
We use an LSTM-based encoder to obtain a representation for the input question $X_i$, which is represented as a matrix in which every row corresponds to the vector representation of each word. 
We use a one-hot vector representation for every word and obtain a word embedding $c_i$ for each word using a Temporal CNN~\cite{zhang_NIPS2015character,Palangi_TASLP2016} module that we parameterize through a function $G(X_i,W_e)$  where $W_e$ are the weights of the temporal CNN. Now this word embedding is fed to an LSTM-based encoder which provides encoding features of the sentence. We use LSTM~\cite{hochreiter_NC1997} 
due to its capability of capturing long term memory~\cite{Palangi_TASLP2016}. As the words are propagated through the network, the network collects more and more semantic information about the sentence. When the network reaches the last word ($L_{th}$ word), the hidden state $h_{L}$ of the network provides a semantic representation of the whole sentence conditioned on all the previously generated words $(q_0,q_1...,q_{t})$. 
Question sentence encoding feature $f_i$ is obtained after passing through an LSTM which is parameterized using the function $F(C_i, W_l)$ where $W_l$ are the weights of the LSTM. This is illustrated in part 1 of Figure~\ref{fig:main_figure}. 

\subsection{Decoder-LSTM}
 The role of decoder is to predict the probability for a whole sentence, given the embedding of input sentence ($f_i$). RNN provides a nice way to condition on previous state value using a fixed length hidden vector. The conditional probability of a sentence token at a particular time step 
 is modeled using an LSTM as used in machine translation~\cite{Sutskever_NIPS2014}. At time step $t$, the conditional probability is denoted by $P( q_{t} | {f_i},q_0,..,q_{t-1})= P( q_{t} | {f_i},h_{t})$, where $h_{t}$ is the hidden state of the LSTM cell at time step $t$.  $h_{t}$ is conditioned on all the previously generated words $(q_0,q_1..,q_{t-1})$ and $q_{t}$ is the next generated word.
 
  Generated question sentence feature $\hat{p_d} = \{\hat{p_1},\dots, \hat{p_T}\}$ is obtained by decoder LSTM which is parameterized using the function $D(f_i, W_{dl})$ where $W_{dl}$ are the weights of the decoder LSTM.
 The output of the word with maximum probability in decoder LSTM cell at step $k$  is input to the LSTM cell at step $k+1$ as shown in Figure~\ref{fig:main_figure}. At $t=-1$, we are feeding the embedding of input sentence obtained by the encoder module. $\hat{Y_i}=\{\hat{q_0},\hat{q_1},...,\hat{q}_{T+1}\}$ are the predicted question tokens for the input $X_i$. Here, we are using $\hat{q_0}$ and $\hat{q}_{T+1}$ as the special START and STOP token respectively.
 The predicted question token ($\hat{q_i}$) is obtained by applying Softmax on the probability distribution $\hat{p_i}$. 
 The question tokens at different time steps are given 
 by the following equations where LSTM refers to the standard LSTM cell equations:
\begin{equation}
 \begin{split}
& d_{-1}=\mbox{Encoder}(f_{i}) \\
& h_0=\mbox{LSTM}(d_{-1})\\
& d_t=W_d*q_t,  \forall t\in \{0,1,2,...T-1\} \\
& {h_{t+1}}=\mbox{LSTM}(d_t,h_{t}), \forall t\in \{0,1,2,...T-1\}\\
& \hat{p}_{t+1} = W_v * h_{t+1} \\
& \hat{q}_{t+1} = \mbox{Softmax}(\hat{p}_{t+1})\\
& \text{Loss}_{t+1}=\text{loss}(\hat{q}_{t+1},q_{t+1})
 \end{split}
\end{equation}
 Where  $\hat{q}_{t+1}$ is the predicted question token and $q_{t+1}$ is the ground truth one. In order to capture local label information, we use the Cross Entropy loss which is given by the following equation:
  \begin{equation}
L_{local}=\frac{-1}{T}\sum_{t=1}^{T} {q_{t} \text{log} \text{P}(\hat{q_{t}}|{q_0,..q_{t-1}})}
\end{equation}
Here $T$ is the total number of sentence tokens, $\text{P}(\hat{q_{t}}|{q_0,..q_{t-1}})$ is the predicted probability of the sentence token, $q_t$ is the ground truth token. 
\subsection{Discriminative-LSTM}
The aim of the Discriminative-LSTM is to make the predicted sentence embedding $f_i^p$ and ground truth sentence embedding $f_i^g$ indistinguishable  as shown in Figure~\ref{fig:main_figure}. Here we pass $\hat{p_d}$ to the shared encoder-LSTM to obtain $f_i^p$ and also the ground truth sentence
to the shared encoder-LSTM to obtain $f_i^g$. The discriminator module estimates a loss function between the generated and ground truth paraphrases. Typically, the discriminator is a binary classifier loss, but here we use a global loss, similar to~\cite{Reed_CVPR2016} which acts on the last hidden state of the recurrent neural network (LSTM).
The main objective of this loss is to bring the generated paraphrase embeddings closer to its ground truth paraphrase embeddings and farther from the other ground truth paraphrase embeddings (other sentences in the batch). Here our discriminator network ensures that the generated embedding can reproduce better paraphrases.
We are using the idea of sharing discriminator parameters with encoder network, to enforce learning of embeddings that not only minimize the local loss (cross entropy), but also the global loss.

Suppose the predicted embeddings of a batch is  $e_p=[f_1^p,f_2^p,.. f_N^p]^T$, where $f_i^p$ is the sentence embedding of $i^{th}$ sentence of the  batch. Similarly ground truth batch embeddings are $e_g=[f_1^g,f_2^g,.. f_N^g]^T$, where $N$ is the batch size, $f_i^p \in \mathds{R}^d$ $f_i^g \in \mathds{R}^d$. The objective of global loss is to maximize the similarity between predicted sentence $f_i^p$ with the ground truth sentence $f_i^g$ of $i^{th}$ sentence and minimize the similarity between $i^{th}$ predicted sentence, $f_i^p$, with $j^{th}$ ground truth sentence, $f_j^g$, in the batch. 
The loss is defined as 
\begin{equation}
L_{global}= \sum_{i=1}^{N}\sum_{j=1}^{N} \max(0,((f_i^p\cdot f_j^g)- (f_i^p \cdot f_i^g)+1))
\end{equation}
Gradient of this loss function is given by 
\begin{equation}
\bigg(\frac{dL}{de_p}\bigg)_i= \sum_{j=1,\\ j\neq i}^{N} (f_j^g-f_i^g)
\end{equation}
\begin{equation}
\bigg(\frac{dL}{de_g}\bigg)_i= \sum_{j=1,\\ j\neq i}^{N} (f_j^p-f_i^p)
\end{equation}

\subsection{Cost function}
\noindent Our objective is to minimize the total loss, that is the sum of local  loss and global loss over all training examples.
  The total loss is: 
  \begin{equation}
    L_{total}= \frac{1}{M} \sum^{M}_{i=1} (L_{local} + L_{global})
\end{equation}
Where $M$ is the total number of examples, $L_{local}$ is the cross entropy loss, $L_{global}$ is the global loss.

\begin{table}[ht] 
\centering
\begin{tabular}{|l|l|l|c|c|c|c|c|}
\hline
\bf Dataset &\bf Model & \bf BLEU1 & \bf BLEU2 &  \bf BLEU3 & \bf BLEU4 &\bf ROUGE  & \bf METEOR \\ \hline
&ED-L(Base Line) &33.7 & 22.3 &18.0 & 12.1 &35.3  &14.3 \\ 
&EDD-G  &40.7 &28.3  &21.1 &16.1  &39.7  &19.6 \\ 
50K&EDD-LG  &40.9 &28.6  &21.3 &16.1  &40.2  &19.8 \\
&EDD-LG(shared) &\textbf{41.1} &\textbf{29.0}  &\textbf{21.5} &\textbf{16.5}  &\textbf{40.6}  &\textbf{20.1}\\ 
\hline
&ED-L(Base Line)  &35.1 & 25.4 &19.6 & 14.4 &37.4  &15.4 \\ 
&EDD-G &42.1 & 29.4 &21.6 & 16.4 &41.4  &20.4 \\ 
100K&EDD-LG &44.2 &31.6  &22.1 &17.9  & 43.6 &22.1 \\ 
&EDD-LG(shared) & \textbf{45.7}& \textbf{32.4} &\textbf{23.8} &\textbf{17.9}  &\textbf{44.9}  &\textbf{23.1}\\ 
\hline
 
\end{tabular}
\caption{\label{score_tab_1}Analysis of variants of our proposed method on Quora Dataset as mentioned in section \ref{ablation_analysis}. Here L and G refer to the Local and Global loss and shared represents the parameter sharing between the discriminator and encoder module. As we can see that our proposed method EDD-LG(shared) clearly outperforms the other ablations on all metrics and detailed analysis is present in section~\ref{ablation_analysis}.}
\end{table}

\section{Experiments}
We perform experiments to better understand the behavior of our proposed embeddings. To achieve this, we benchmark Encoder Decoder Discriminator Local-Global (shared) (EDD-LG(shared)) embeddings on two text understanding problems, Paraphrase Generation and Sentiment Analysis. We use the Quora question pairs dataset \footnote{website: \url{https://data.quora.com/First-Quora-Dataset-Release-Question-Pairs}} for paraphrase generation and Stanford Sentiment Treebank dataset~\cite{Socher_EMNLP2013} for sentiment analysis. In this section we describe the different datasets, experimental setup and results of our experiments. 

\subsection{Paraphrase Generation}
Paraphrase generation is an important problem in many NLP applications such as question answering, information retrieval, information extraction, and summarization. It involves generation of similar meaning sentences. 

\subsubsection{Dataset}
We use the newly released Quora question pairs dataset for this task. It consists of over 400K potential question duplicate pairs.
As pointed out in~\cite{gupta_AAAI2017}, the question pairs having the binary value 1 are the ones which are actually the paraphrase of each other and the others are duplicate questions. 
So, we choose all such question pairs with binary value 1. There are a total of 149K such questions. 
Some examples of generated question-paraphrase pairs are provided in Table~\ref{tab:paraphrase_samples}. More results are present in the appendix.

\begin{table}[ht]
\centering
\begin{tabular}{|l|l|l|c|c|c|c|c|}
\hline \bf Dataset &\bf Model & \bf BLEU1 & \bf METEOR & \bf TER \\ \hline 
&Unsupervised VAE~\cite{gupta_AAAI2017} & 8.3 &12.2  &83.7\\ 
&VAE-S~\cite{gupta_AAAI2017}  &11.9 &17.4  &69.4\\
50K&VAE-SVG~\cite{gupta_AAAI2017} &17.1 & 21.3 &63.1 \\
&VAE-SVG-eq~\cite{gupta_AAAI2017} &17.4 & \textbf{21.4} & 61.9\\ 
&EDD-G (\textbf{Ours}) &40.7 & 19.7 & 51.2\\
&EDD-LG(\textbf{Ours}) &40.9 & 19.8 & 51.0\\
&EDD-LG(shared)(\textbf{Ours}) &\textbf{41.1} & 20.1 & \textbf{50.8}\\
\hline
&Unsupervised~\cite{gupta_AAAI2017} & 10.6&14.3  &79.9 \\ 
&VAE-S~\cite{gupta_AAAI2017}  & 17.5& 21.6 &67.1\\
100K&VAE-SVG~\cite{gupta_AAAI2017} &22.5 & 24.6 &55.7\\
&VAE-SVG-eq~\cite{gupta_AAAI2017} &22.9 & \textbf{24.7} &55.0 \\
&EDD-G (\textbf{Ours}) &42.1 & 20.4 & 49.9\\
&EDD-LG(\textbf{Ours}) &44.2 & 22.1 & 48.3\\
&EDD-LG(shared)(\textbf{Ours}) &\textbf{45.7} &23.1 & \textbf{47.5}\\
\hline
\end{tabular}
\caption{\label{score_tab_2}Analysis of Baselines and State-of-the-Art methods for paraphrase generation on Quora dataset. As we can see clearly that our model outperforms the state-of-the-art methods by a significant margin in terms of BLEU and TER scores. Detailed analysis is present in section~\ref{sec:Baselines_analysis}. A lower TER score is better whereas for the other metrics, a higher score is better. Details for the metrics are present in the appendix.}
\end{table}
\subsubsection{Experimental Protocols} We follow the experimental protocols mentioned in~\cite{gupta_AAAI2017} for the Quora Question Pairs dataset. In our experiments, we divide the dataset into 2 parts 145K and 4K question pairs. We use these as our training and testing sets. We further divide the training set into 50K and 100K dataset sizes and use the rest 45K as our validation set. We also followed the dataset split mentioned in~\cite{li_arXiv2017Paraphrase} to calculate the accuracies on a different test set and provide the results on our project webpage.
We  trained our model end-to-end using local loss (cross entropy loss) and global loss. We have used RMSPROP  optimizer to update the model parameter and found these hyperparameter values to work best to train the Paraphrase Generation Network: $\text{learning rate} =0.0008, \text{batch size} = 150, \alpha = 0.99,  \epsilon=1e-8$.
We have used learning rate decay to decrease the learning rate on every epoch by a factor given by:
\[\text{Decay\_factor}=exp\left(\frac{log(0.1)}{a*b} \right)\] where $a=1500$ and $b=1250$ are set empirically.
\begin{table}[ht]

\centering
\begin{tabular}{|>{\arraybackslash}m{0.035\textwidth}|>{\arraybackslash}m{0.25\textwidth}|>{\arraybackslash}m{0.3\textwidth}|>{\arraybackslash}m{0.25\textwidth}|}
\hline S.No&\bf Original Question &\bf Ground Truth Paraphrase & \bf Generated Paraphrase \\ \hline 
1&Is university really worth it?	&Is college even worth it?	&Is college really worth it? \\ \hline
2&Why India is against CPEC? &Why does India oppose CPEC? &Why India is against Pakistan? \\ \hline
3&How can I find investors for my tech startup?	&How can I find investors for my startup on Quora?	&How can I find investors for my startup business? \\\hline
4&What is your view/opinion about surgical strike by the Indian Army?&	What world nations think about the surgical strike on POK launch pads and what is the reaction of Pakistan?&	What is your opinion about the surgical strike on Kashmir like? \\ \hline
5&What will be Hillary Clinton's strategy for India if she becomes US President? & What would be Hillary Clinton's foreign policy towards India if elected as the President of United States?	& What will be Hillary Clinton's policy towards India if she becomes president? \\ \hline
\end{tabular}
\caption{\label{tab:paraphrase_samples}Examples of Paraphrase generation on Quora Dataset. We observe that our model is able to understand abbreviations as well and then ask questions on the basis of that as is the case in the second example.}
\end{table}
\subsubsection{Ablation Analysis}\label{ablation_analysis}
We experimented with different variations for our proposed method. We start with baseline model  which we take as a simple encoder and decoder network with only the local loss (ED-Local)~\cite{Sutskever_NIPS2014}. Further we have experimented with encoder-decoder and a discriminator network with only global loss (EDD-Global) to distinguish the ground truth paraphrase with the predicted one. Another variation of our model is used both the global and local loss (EDD-LG). The discriminator is the same as our proposed method, only the weight sharing is absent in this case.
Finally, we make the discriminator share weights with the encoder and train this network with both the losses (EDD-LG(shared)). The analyses are given in table~\ref{score_tab_1}. Among the ablations, the proposed EDD-LG(shared) method works way better than the other variants in terms of BLEU and METEOR metrics by achieving an improvement of 8\% and 6\% in the scores respectively over the baseline method for 50K dataset and an improvement of 10\% and 7\% in the scores respectively for 100K dataset.
\subsubsection{Baseline and State-of-the-Art Method Analysis}\label{sec:Baselines_analysis}
There has been relatively less work on this dataset and the only work which we came across was that of~\cite{gupta_AAAI2017}. 
We further compare our method EDD-LG(shared) model with their VAE-SVG-eq which is the current state-of-the-art on Quora datset. Also we provide comparisons with other methods proposed by them in table~\ref{score_tab_2}. 
As we can see from the table that we achieve a significant improvement of 24\% in BLEU score and 11\% in TER score (A lower TER score is better) for 50K dataset and similarly 22\% in BLEU score and 7.5\% in TER score for 100K dataset.

\subsubsection{Statistical Significance Analysis} 
\begin{figure}[ht]
	\centering
	\includegraphics[width=0.45\textwidth]{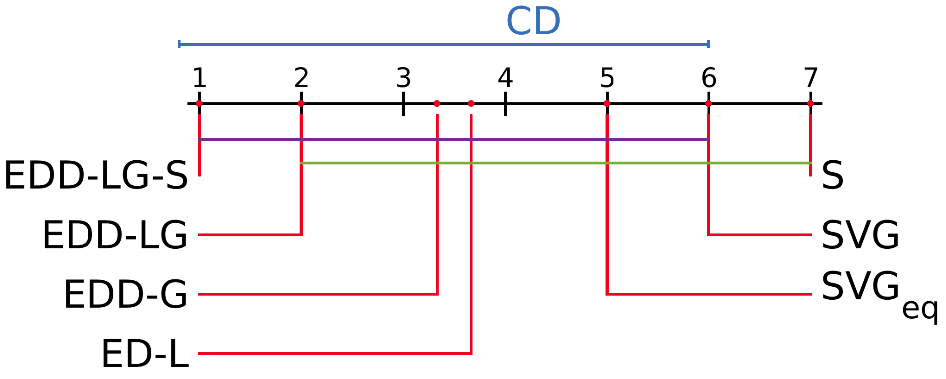}
	\caption{The mean rank of all the models on the basis of BLEU score are plotted on the x-axis. Here EDD-LG-S refers to our EDD-LG shared model and others are the different variations of our model described in section \ref{ablation_analysis} and the models on the right are the different variations proposed in~\cite{gupta_AAAI2017}. Also the colored lines between the two models represents that these models are not significantly different from each other. CD=5.199,p=0.0069}
	\label{fig:result_1_B}
\end{figure}

We have analysed statistical significance~\cite{Demvsar_JMLR2006} for our proposed embeddings against different ablations and the state-of-the-art methods for the paraphrase generation task. 
The Critical Difference (CD) for Nemenyi~\cite{Fivser_PLOS2016} test depends upon the given $\alpha$ (confidence level, which is 0.05 in our case) for average ranks and N (number of tested datasets). If the difference in the rank of the two methods lies within CD, then they are not significantly different, otherwise they are statistically different. Figure ~\ref{fig:result_1_B} visualizes the post hoc analysis using the CD diagram. From the figure, it is clear that our embeddings work best and the results are significantly different from the state-of-the-art methods.

\begin{table}[ht]
\centering
\begin{tabular}{|>{\arraybackslash}m{0.55\textwidth}|>{\centering\arraybackslash}m{0.2\textwidth}|>{\centering\arraybackslash}m{0.2\textwidth}|}
\hline {\bf Model}  & \bf Error Rate (Fine-Grained) \\\hline 
Naive Bayes~\cite{Socher_EMNLP2013}  & 59.0  \\
SVMs~\cite{Socher_EMNLP2013}  & 59.3 \\ 
Bigram Naive Bayes~\cite{Socher_EMNLP2013}   & 58.1  \\
Word Vector Averaging~\cite{Socher_EMNLP2013}   & 67.3\\
Recursive Neural Network~\cite{Socher_EMNLP2013}  & 56.8  \\
Matrix Vector-RNN~\cite{Socher_EMNLP2013}  & 55.6  \\
Recursive Neural Tensor Network~\cite{Socher_EMNLP2013}  & 54.3  \\
Paragraph Vector~\cite{Le_ICML2014}   & {51.3}\\ \hline
EDD-LG(shared) (\bf Ours) & \bf 35.6 \\
\hline
\end{tabular}
\caption{\label{score_tab_4}Performance of our method compared to other approaches on the Stanford Sentiment Treebank Dataset. The error rates of other methods are reported in~\cite{Le_ICML2014}}
\end{table}

\subsection{Sentiment Analysis with Stanford Sentiment Treebank (SST) Dataset}


\subsubsection{Dataset}
This dataset consists of sentiment labels for different movie reviews and was first proposed by~\cite{Pang_ACL2005}.~\cite{Socher_EMNLP2013} extended this by parsing the reviews to subphrases and then fine-graining the sentiment labels for all the phrases of movies reviews using Amazon Mechanical Turk. The labels are classified into 5 sentiment classes, namely \{Very Negative, Negative, Neutral, Positive,
Very Positive\}.  This dataset contains a total 126k phrases for training set, 30k phrases for validation set and 66k phrases for test set.

\subsubsection{Tasks and Baselines} In~\cite{Socher_EMNLP2013}, the authors
propose two ways of benchmarking. We consider the 5-way fine-grained classification task where
the labels are \{Very Negative, Negative, Neutral, Positive,
Very Positive\}. The
other axis of variation is in terms of whether we should label
the entire sentence or all phrases in the sentence. In this
work we only consider labeling all the phrases.~\cite{Socher_EMNLP2013} apply several methods
to this dataset and we show their performance in table~\ref{score_tab_4}.

\begin{table}[ht]
\small
\centering
\begin{tabular}{|>{\arraybackslash}m{0.075\textwidth}|>{\arraybackslash}m{0.73\textwidth}|m{0.11\textwidth}|}
\hline \bf Phrase ID &\bf Phrase & \bf Sentiment \\ \hline 


162970&	The heaviest, most joyless movie &\\
159901&	Even by dumb action-movie standards, Ballistic : Ecks vs. Sever is a dumb action movie. &\\
158280&	Nonsensical, dull ``cyber-horror'' flick is a grim, hollow exercise in flat scares and bad acting &Very Negative\\
159050&	This one is pretty miserable, resorting to string-pulling rather than legitimate character development and intelligent plotting. &\\
157130&	The most hopelessly monotonous film of the year, noteworthy only for the gimmick of being filmed as a single unbroken 87-minute take. &\\
\hline
156368&	No good jokes, no good scenes, barely a moment &\\
157880&	Although it bangs a very cliched drum at times &\\
159269&	They take a long time to get to its gasp-inducing ending. & Negative\\
157144&	Noteworthy only for the gimmick of being filmed as a single unbroken 87-minute &\\
156869&	Done a great disservice by a lack of critical distance and a sad trust in liberal arts college bumper sticker platitudes&\\ \hline
221765&	A hero can stumble sometimes. &\\
222069&	Spiritual rebirth to bruising defeat&\\
218959&	An examination of a society in transition &Neutral\\
221444&	A country still dealing with its fascist past& \\
156757&	Have to know about music to appreciate the film's easygoing blend of comedy and romance & \\
\hline

157663&	A wildly funny prison caper. &\\
157850&	This is a movie that's got oodles of style and substance. & \\
157879&	Although it bangs a very cliched drum at times, this crowd-pleaser's fresh dialogue, energetic music, and good-natured spunk are often infectious. & Positive\\
156756&	You don't have to know about music to appreciate the film's easygoing blend of comedy and romance. &\\
157382&	Though of particular interest to students and enthusiast of international dance and world music, the film is designed to make viewers of all ages, cultural backgrounds and rhythmic ability want to get up and dance. &\\
\hline
162398&	A comic gem with some serious sparkles. &\\
156238&	Delivers a performance of striking skill and depth &\\
157290&	What Jackson has accomplished here is amazing on a technical level. &Very Positive\\
160925&	A historical epic with the courage of its convictions about both scope and detail. &\\
161048&	This warm and gentle romantic comedy has enough interesting characters to fill several movies, and its ample charms should win over the most hard-hearted cynics.&\\ 
\hline
\end{tabular}

\caption{\label{score_tab_sup}Examples of Sentiment classification on test set of kaggle competition dataset.}
\end{table}

\subsubsection{Experimental Protocols} 
For the task of Sentiment analysis, we are using a similar method of performing the experiments as used by~\cite{Socher_EMNLP2013}.
We treat every subphrase in the dataset as a separate sentence and learn their corresponding representations. We then feed these to a logistic regression to predict the movie ratings.
During inference time, we used a method simialr to~\cite{Le_ICML2014} in which we freeze the representation of every word and use this to construct a representation for the test sentences which are then fed to a logistic regression for predicting the ratings. 
In order to train a sentiment classification model, we have used RMSPROP, to  optimize the classification model parameter and we found these hyperparameter values to be working best for our case: $\text{learning rate} =0.00009, \text{batch size} = 200, \alpha = 0.9,  \epsilon=1e-8$.\\
\subsubsection{Results}
We report the error rates of different methods in table~\ref{score_tab_4}. We can clearly see that the performance of bag-of-words or bag-of-n-grams models (the first four models in the table) is not up to the mark and instead the advanced methods (such as Recursive Neural Network~\cite{Socher_EMNLP2013}) perform better on sentiment analysis task. Our method outperforms all these methods by an absolute margin of 15.7\% which is a significant increase considering the rate of progress on this task. We have also uploaded our models to the online competition on Rotten Tomatoes dataset \footnote{website: \url{www.kaggle.com/c/sentiment-analysis-on-movie-reviews}} and obtained an accuracy of 62.606\% on their test-set of 66K phrases. \\
We provide 5 examples for each sentiment in table~\ref{score_tab_sup}. We can see clearly that our proposed embeddings are able to get the complete meaning of smaller as well as larger sentences. For example, our model classifies `Although it bangs a very cliched drum at times' as Negative and `Although it bangs a very cliched drum at times, this crowd-pleaser's fresh dialogue, energetic music, and good-natured spunk are often infectious.' as positive showing that it is able to understand the finer details of language. More results and visualisations showing the part of the phrase to which the model attends while classifying are present in the appendix. 
The link for the project website and code is provided here \footnote{Project website: \url{https://badripatro.github.io/Question-Paraphrases/}}.
\section{Conclusion}

In this paper we have proposed a sentence embedding using a sequential encoder-decoder with a pairwise discriminator. We have experimented with this text embedding method for paraphrase generation and sentiment analysis. We also provided experimental analysis which justifies that a pairwise discriminator outperforms the previous state-of-art methods for NLP tasks. We also performed ablation analysis for our method, and our method outperforms all of them in terms of BLEU, METEOR and TER scores. We plan to generalize this to other text understanding tasks and also extend the same idea in vision domain.

\bibliographystyle{acl}
\bibliography{coling2018}

\appendix

\newpage
 \section{Appendix}
\subsection{Quantitative Evaluation}

We use automatic evaluation metrics which are prevalent in machine translation domain: BLEU~\cite{Papineni_ACL2002}, METEOR~\cite{Banerjee_ACL2005}, ROUGE-n~\cite{Lin_ACL2004} and Translation Error Rate (TER) \cite{Snover_AMT2006}. 
These metrics perform well for Paraphrase generation task and also have a higher correlation with human judgments~\cite{Madnani_ACL2012,Wubben_LOT2010}. 
BLEU uses n-gram precision between the ground truth and the predicted paraphrase. considers exact match between reference
whereas ROUGE considers recall for the same. On the other hand, METEOR uses stemming and synonyms (using WordNet) and is based on the harmonic mean of unigram-precision and unigram-recall.
TER is based on the number of edits (insertions, deletions, substitutions, shifts) required to convert the generated output into the ground truth paraphrases and quite obviously a lower TER score is better whereas other metrics prefer a higher score for showing improved performance. We provided our results using all these metrics and compared it with existing baselines.

\subsection{Paraphrase Generation}
Here we provide some more examples of the paraphrase generation task in table~\ref{tab:paraphrase_samples1}. Our model is also able to generate sentences which capture higher level semantics like in the last example of table~\ref{tab:paraphrase_samples1}. 

\begin{table}[ht]

\centering
\begin{tabular}{|>{\arraybackslash}m{0.035\textwidth}|>{\arraybackslash}m{0.25\textwidth}|>{\arraybackslash}m{0.3\textwidth}|>{\arraybackslash}m{0.3\textwidth}|}
\hline S.No &\bf Original Question &\bf Ground Truth Paraphrase & \bf Generated Paraphrase \\ \hline 
1&How do I add content on Quora?	&How do I add content under a title at Quora?	&How do I add images on Quora ? \\ \hline
2& Is it possible to get a long distance ex back?  & Long distance relationship: How to win my ex-gf back? &Is it possible to get a long distance relationship back ? \\ \hline
3&How many countries are there in the world? Thanks! 	&  How many countries are there in total?  	&How many countries are there in the world ? What are they ? \\\hline
4&What is the reason behind abrupt removal of Cyrus Mistry?&	Why did the Tata Sons sacked Cyrus Mistry?&	 What is the reason behind firing of Cyrus Mistry ? \\ \hline
5&  What are some extremely early signs of pregnancy? &  What are the common first signs of pregnancy?  How can I tell if I'm pregnant? What are the symptoms?	& What are some early signs of pregnancy ? \\ \hline

6&  How can I improve my critical reading skills?  &   What are some ways to improve critical reading and reading comprehension skills? 	& How can I improve my presence of mind ? \\ \hline
\end{tabular}
\caption{\label{tab:paraphrase_samples1}Examples of Paraphrase generation on Quora Dataset.}
\end{table}
\subsection{Sentiment Analysis}
We also provide visualization of different parts of the sentence on which our model focuses while predicting the sentiment in Figure~\ref{fig:visualize} and some more examples of the Sentiment analysis task on SST dataset in Table~\ref{score_tab_sup1}. 
\begin{table}[ht]
\small
\centering
\begin{tabular}{|>{\arraybackslash}m{0.05\textwidth}|>{\arraybackslash}m{0.78\textwidth}|m{0.09\textwidth}|}
\hline \bf Phrase ID &\bf Phrase & \bf Sentiment \\ \hline 


156628&	The movie is just a plain old monster&\\ 
157078&	a really bad community theater production of West Side Story &\\
159749&	Suffers from rambling , repetitive dialogue and the visual drabness endemic to digital video . &\\
163425&	The picture , scored by a perversely cheerful Marcus Miller accordion\/harmonica\/banjo abomination , is a monument to bad in all its florid variety . &\\
163483&	lapses quite casually into the absurd & Very Negative\\
163882&	It all drags on so interminably it 's like watching a miserable relationship unfold in real time . &\\
164436&	Your film becomes boring , and your dialogue is n't smart &\\
165179&	Another big , dumb action movie in the vein of XXX , The Transporter is riddled with plot holes big enough for its titular hero to drive his sleek black BMW through . &\\\hline
156567&	It would be hard to think of a recent movie that has worked this hard to achieve this little fun&\\ 
156689&	A depressing confirmation &\\
157730&	There 's not enough here to justify the almost two hours. &\\
157695&	a snapshot of a dangerous political situation on the verge of coming to a head &\\
158814&	It is ridiculous , of course &Negative\\
159281&	A mostly tired retread of several other mob tales.&\\
159632&	We are left with a superficial snapshot that , however engaging , is insufficiently enlightening and inviting .&\\ 
159770&	It 's as flat as an open can of pop left sitting in the sun .  &\\\hline

156890&	liberal arts college bumper sticker platitudes & \\
160247&	the movie 's power as a work of drama &\\
160754&	Schweig , who carries the film on his broad , handsome shoulders &\\
160773&	to hope for any chance of enjoying this film &\\
201255&	also examining its significance for those who take part &Neutral\\
201371&	those who like long books and movies &\\
221444&	a country still dealing with its fascist past&\\
222102&	used to come along for an integral part of the ride &\\
\hline

157441&	the film is packed with information and impressions . &\\
157879&	Although it bangs a very cliched drum at times , this crowd-pleaser 's fresh dialogue , energetic music , and good-natured spunk are often infectious. &\\
157663&	A wildly funny prison caper. &\\
157749&	This is one for the ages. &Positive\\
157806&	George Clooney proves he 's quite a talented director and Sam Rockwell shows us he 's a world-class actor with Confessions of a Dangerous Mind . &\\
157850&	this is a movie that 's got oodles of style and substance . &\\\hline


157742&	Kinnear gives a tremendous performance . &\\
160562&	The film is painfully authentic , and the performances of the young players are utterly convincing .&\\ 
160925&	A historical epic with the courage of its convictions about both scope and detail. &\\
161048&	This warm and gentle romantic comedy has enough interesting characters to fill several movies , and its ample charms should win over the most hard-hearted cynics .&Very Positive\\ 
161459&	is engrossing and moving in its own right & \\
162398&	A comic gem with some serious sparkles . &\\
162779&	a sophisticated , funny and good-natured treat , slight but a pleasure &\\
163228&	Khouri then gets terrific performances from them all .&\\\hline


\end{tabular}
\caption{\label{score_tab_sup1}Examples of Sentiment classification on test set of kaggle dataset.}
\end{table}

\subsubsection{Sentiment Visualization of the sentence}
\begin{figure*}[ht]
\centering
\includegraphics[width=1.0\columnwidth]{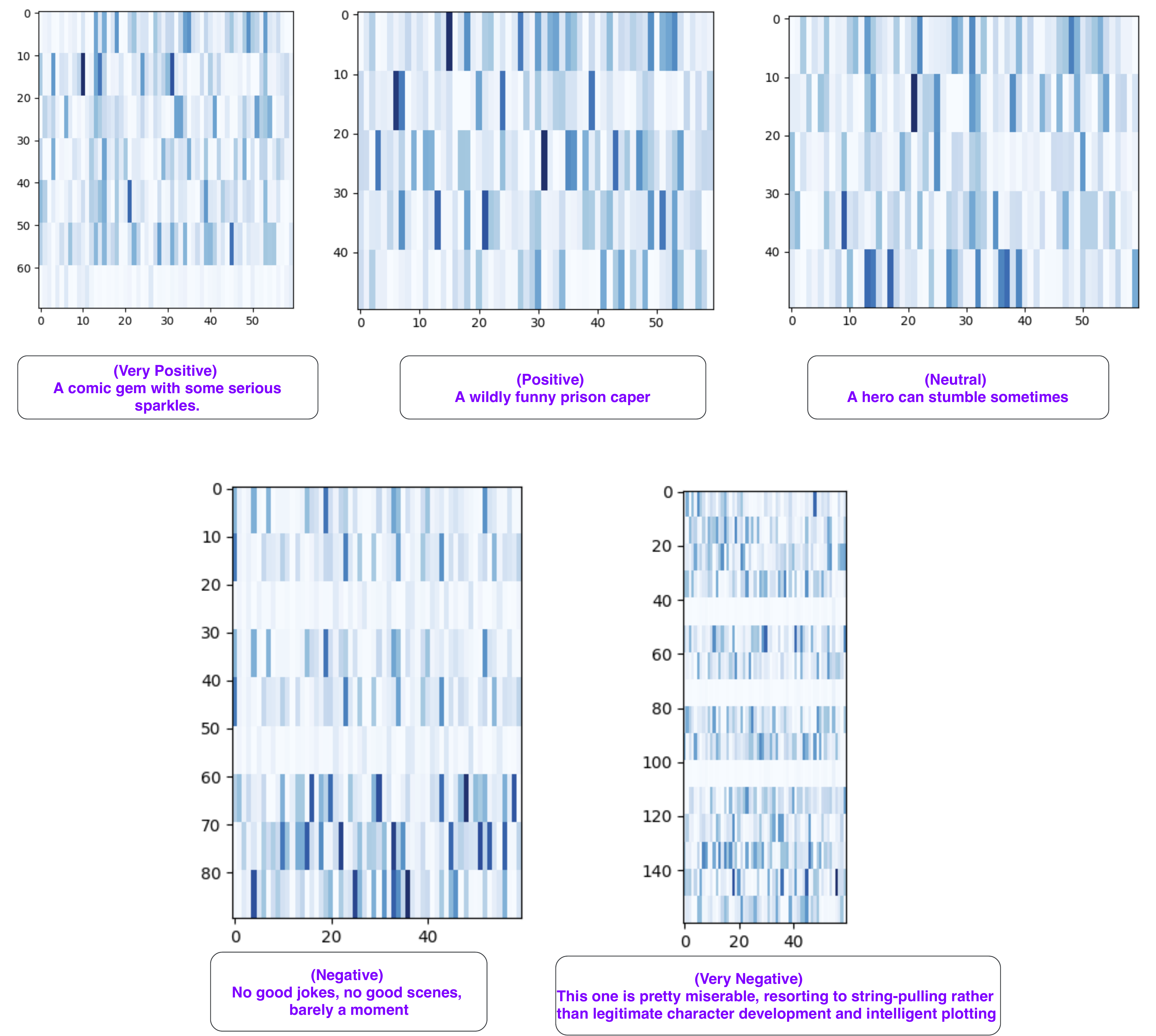} 
\caption{These are the visualisations for the sentiment analysis for some examples and we can clearly see that our model focuses on those words which we humans focus while deciding the sentiment for any sentence. In the second image, `wildly' and `funny' are emphasised more than the other words. }
\label{fig:visualize}
\end{figure*}

\cite{li_NAACL2016visualizing} have proposed a mechanism to visualize language features. We conducted a toy experiment for our EDD-LG(shared) model. Figure~\ref{fig:visualize} represents saliency heat map for EDD-LG(shared) model sentiment analysis. We obtained 60 dimensional feature maps for each word present in the target sentence. The heat map captures the measure of influence of the sentimental decision.
In the heat map, each word of a sentence (from top to bottom, first word at top) represents its contribution for making the sentimental decision.  
For example in the first image in~\ref{fig:visualize}, the word `comic' contributed more (2nd word, row 10-20).
Similarly in the second image, first, second, and third (`A',`wildly',`funny') words have more influence for making this sentence have a positive sentiment. 

\end{document}